\documentclass[conference]{IEEEtran}
\usepackage{times}

\usepackage[numbers]{natbib}
\usepackage{multicol}
\usepackage[bookmarks=true]{hyperref}
\usepackage{amsmath} 
\usepackage{bm}
\usepackage{graphicx}
\usepackage{subfigure}
\usepackage{amsfonts}
\usepackage{algorithm}
\usepackage{algorithmicx}
\usepackage{algpseudocode}
\usepackage{flushend}
\newcommand{\mystep}[1]{{\vspace{2.5mm}\noindent\textbf{#1}}}

\begin{document}

\title{Passive Static Equilibrium with Frictional
  Contacts and Application to Grasp Stability Analysis}


\author{
  \authorblockN{
    Maximilian Haas-Heger\authorrefmark{1},
    Christos Papadimitriou\authorrefmark{2},
    Mihalis Yannakakis\authorrefmark{3}, 
    Garud Iyengar\authorrefmark{4} and
    Matei Ciocarlie\authorrefmark{5}}
  \authorblockA{
    \authorrefmark{1}Department of Mechanical Engineering; email:  m.haas@columbia.edu\\
    \authorrefmark{2}Department of Computer Science; email: christos@cs.columbia.edu\\
    \authorrefmark{3}Department of Computer Science; email: mihalis@cs.columbia.edu\\
    \authorrefmark{4}Department of Industrial Engineering and Operations Research; email: garud@ieor.columbia.edu\\
    \authorrefmark{5}Department of Mechanical Engineering; email: matei.ciocarlie@columbia.edu\\
  }
  \authorblockA{Columbia University, New York, NY 10027}
}

%

\maketitle

\begin{abstract}
This paper studies the problem of passive grasp stability under an
external disturbance, that is, the ability of a grasp to resist a
disturbance through passive responses at the contacts. To obtain
physically consistent results, such a model must account for friction
phenomena at each contact; the difficulty is that friction forces
depend in non-linear fashion on contact behavior (stick or slip). We
develop the first polynomial-time algorithm which either solves such
complex equilibrium constraints for two-dimensional grasps, or
otherwise concludes that no solution exists. To achieve this, we show
that the number of possible ``slip states'' (where each contact is
labeled as either sticking or slipping) that must be considered is
polynomial (in fact quadratic) in the number of contacts, and not
exponential as previously thought. Our algorithm captures passive
response behaviors at each contact, while accounting for constraints
on friction forces such as the maximum dissipation principle.
\end{abstract}

\IEEEpeerreviewmaketitle

\section{INTRODUCTION}

Stability analysis in the presence of frictional contacts is one of
the foundational problems of multi-fingered robotic manipulation. In
particular, a key characteristic of a grasp is the ability to resist
given disturbances, formulated as wrenches externally applied to the
grasped object. This is equivalent to determining the stability of a
multi-body system under applied loads, a problem that is pervasive in
grasp analysis but also encountered in other scenarios, including
simulation of general rigid bodies with frictional contacts.

\begin{figure}[t]
  \centering
  \includegraphics[width=0.95\linewidth]{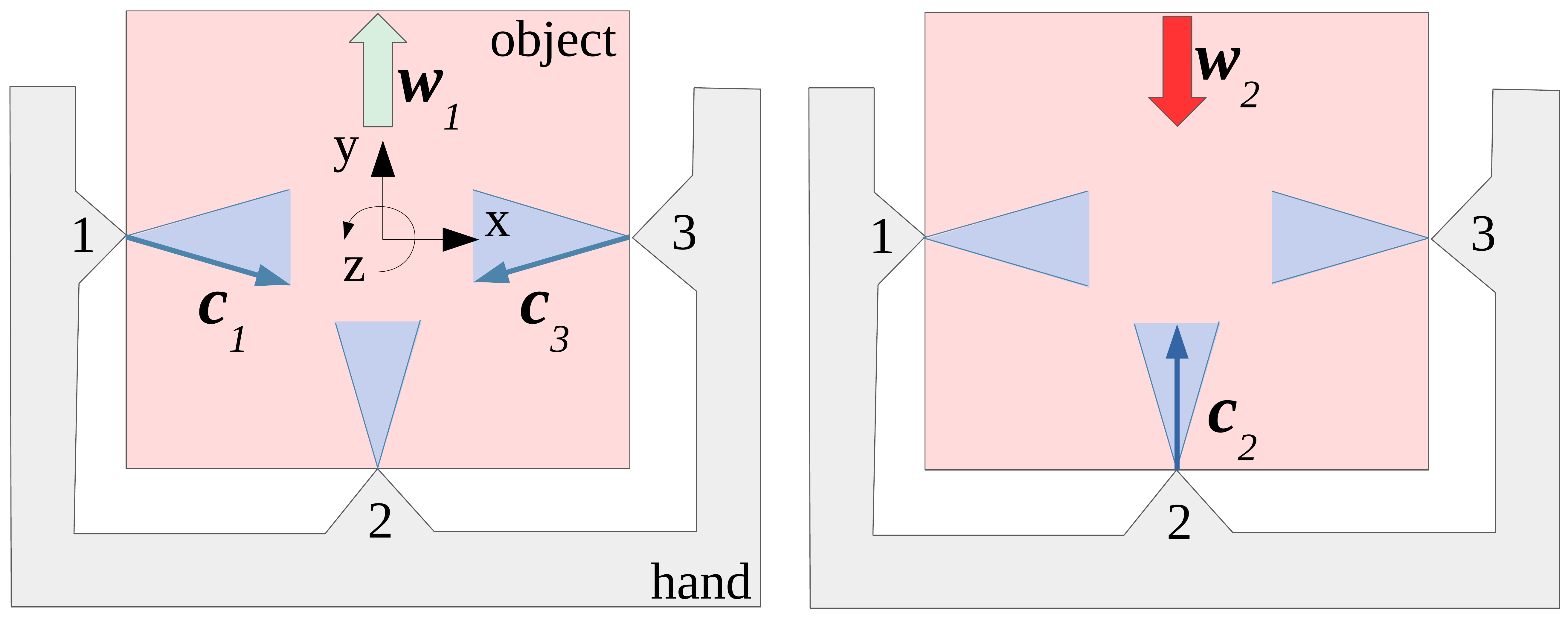} 
  \caption{A grasping scenario where a hand establishes multiple
    frictional contacts (numbered 1-3) with a target object. External
    disturbance $\bm{w}_1$ (left, pushing the object up) can be
    resisted by contact forces $\bm{c}_1$ and $\bm{c}_3$, but only if
    contacts 1 and 3 have been actively pre-loaded with enough normal
    force to generate the corresponding friction forces. In contrast,
    disturbance $\bm{w}_2$ (right, pushing the object down),
    regardless of its magnitude, will always be passively resisted by
    contact force $\bm{c}_2$.}
  \label{fig:grasp}
\end{figure}

We believe that an important distinction to make in such analysis is
between active and passive stability. Consider for example the grasp
shown in Fig.~\ref{fig:grasp}. Will the grasp resist each of the two
disturbances $\bm{w}_1,\bm{w}_2$ applied to the object? Clearly, in
each case, there exists a combination of contact forces that obey
friction laws (are inside their respective friction cones) and sum up
to counterbalance the disturbance ($\bm{c}_1$ together with $\bm{c_3}$
in the case of $\bm{w}_1$, and $\bm{c_2}$ in the case of
$\bm{w}_2$). However, intuition tells us that $\bm{c}_1$ and
$\bm{c_3}$ can \textit{only} arise if contacts 1 and 3 have been
preloaded with enough normal force to sustain the needed level of
friction. In the absence of such a preload, or if the magnitude of the
disturbance is too large relative to the preload, the object will just
slip out. In contrast, $\bm{c}_2$ arises passively, strictly in
response to the disturbance, and matching it in magnitude, thus
$\bm{w}_2$ is \textit{always} resisted.

To account for these differences, we believe that a grasp stability
model should be able to answer the following query, which we refer to
as passive grasp stability. We assume that we are given the geometry
of the grasp (i.e. the shape of the hand and object, and thus the
contact locations) as well as the contact preloads actively applied by
the motors (if any). A known external wrench is then applied to the
object. Our query is formulated as: \textit{Based only on passive
  effects (arising naturally in response to the disturbance), will the
  system find a way to rebalance contact forces allowing it to remain
  in static equilibrium, or will the application of the disturbance
  lead to movement?} We seek a model that provides {\em global
  guarantees:} if no solution to the equilibrium problem is found, we
should be able to guarantee that none exists for the given instance.

Despite significant efforts (which we review in the next section), no existing
grasp model achieves all these goals. Numerous studies do not consider contact
preload when assessing grasp stability, or do not make the distinction between
active and passive effects. Other approaches use simplified models for
determining contact responses to external forces and fail to account for
constraints arising from the dissipation of energy through friction. Finally,
previous work searches for equilibrium configurations using iterative
algorithms that run the risk of getting stuck in local optima, and thus cannot
make strong global guarantees about the absence of a solution.

In this paper, we introduce a grasp stability model that accounts for
all these constraints, and, in the two-dimensional case, allows for an
efficient (polynomial time) algorithm to answer the passive stability
query. We use more realistic friction constraints, where frictional
forces are not linearly related to the externally applied wrench, and
obey the maximum dissipation principle. Our main contributions are as
follows:
\begin{itemize}
  \item For two-dimensional grasps, we show that the total number of
    possible ``slip states'' for the system (where each contact is
    labeled as sticking or slipping, and the direction of slip is also
    determined) that must be considered under rigid body constraints
    is polynomial --- in fact, quadratic --- in the number of contacts.
  \item We use this result to derive the first polynomial-time
    algorithm that can provably determine if a solution exists to the
    passive equilibrium problem for a multi-contact grasp under an
    externally applied wrench, using friction constraints that obey
    the maximum dissipation principle.
\end{itemize}
The ability to make strong guarantees about the existence of
equilibrium is one of the most attractive features of analytical
models as compared to data-driven approaches. Previous methods
either sacrifice this ability (e.g. by introducing iterative
algorithms that are not guaranteed to converge), or simplify the
constraints such that results can be physically
inconsistent. We attempt to avoid such a trade off here, while
preserving the ability to solve the equilibrium query efficiently.

\section{RELATED WORK}

One class of trivially stable robotic grasps consists of form-closure grasps,
which completely immobilize the grasped object. As the object is completely
kinematically constrained, it will be stable in the face of arbitrary applied
wrenches - even in the absence of friction. Checking if a grasp has form
closure can be done by solving a linear program~\cite{YUNHUI99}. However, form
closure grasps require a large number of contacts (at least 7 in the three
dimensional case~\cite{SOMOV1900},\cite{CHRISTOS90}) and are not generally achievable with most
robotic hands. Hence, grasps that do not exhibit form closure are of primary
interest to this investigation. As was described in the Introduction, the
existence of a solution to the equilibrium equations alone is not a sufficient
condition for stability. Neglecting the mechanisms by which contact wrenches
arise leads to false positives in cases where solutions exist, but do not
arise in practice.

For a given robotic grasp, Pang et al.~\cite{ZAMM643} group applied
external wrenches into three classes:
\begin{itemize}
\item Weakly Stable Loads: a solution to the equilibrium equations
exists for given friction coefficients at the contacts. This can be
tested for by solving a linear program.
\item Strongly Stable Loads: A subset of Weakly Stable Loads, the
applied load leads to zero workpiece acceleration for given friction
coefficients. This is equivalent to non-positive virtual work for every
virtual motion that satisfies the kinematic constraints of the grasp.
\item Frictionless Stable Loads: A small subset of the Strongly Stable
Loads. Pang et al. show that if a load is weakly or strongly stable in
the frictionless case, it is also strongly stable for all positive
friction coefficients. Membership can be tested for by solving a
linear program.
\end{itemize}
An algorithm very commonly used in practice to determine the total space of
possible resultant wrenches as long as each individual contact force obeys
(linearized) friction constraints was introduced by Ferrari and
Canny~\cite{FERRARI92}. They call this space the Grasp Wrench Space (GWS),
which can also be used to test a grasp for force closure. This geometric method
is an example of an algorithm we can use to determine the weakly stable loads
as described above. It is an example of a method for stability analysis
that does not account for the distinction between active and passive wrench
reaction and the necessity of a preload in a grasp to withstand certain
applied wrenches. In general - when trying to determine the stability of a
grasp - an enumeration of the first class of applied loads is of limited use
due to the possibility of false positives.

The final class of loads as defined by Pang and summarized above is overly
conservative, as friction is a powerful tool to achieve stable grasps. The
grasp in Fig.~\ref{fig:grasp} for example is unstable in the frictionless
case. The second class of loads is most useful for grasp stability analysis.
Given a load known to be in the first class, in order to test for membership
in the second we need to develop an algorithm that tests if the physically
correct solution (i.e. the one that would arise in reality) lies in the space
of solutions to the equilibrium equations; we need to develop further
constraints that discriminate if a solution to the equilibrium equations is
physically correct.

One approach to this problem is to resolve the statical indeterminacy
by introduction of compliance. This reduces the solution space to at
most a single solution, which is deemed to be the physically correct
one. Of course, the physical correctness depends on the correctness of
the assumptions made in resolving the statical indeterminacy. In his
works~\cite{BICCHI93}\cite{BICCHI94}\cite{BICCHI95} Bicchi assumes a linear
compliance matrix (see also~\cite{CUTKOSKY_COMPLIANCE}), which is also
a potential limitation, as it assumes a linear stiffness of the
contacts. Friction forces, however, are non linear with respect to the
relative sliding motion at the contacts.

Prattichizzo et al.~\cite{PRATTICHIZZO97} built on the previous work by Bicchi
and developed tools for grasp force optimization that specifically take into
account the kinematics of the hand. They also improve on the linear friction
model used by Bicchi in order to alleviate some limitations of the linear
model we will discuss in the next section. The complexity of their algorithm
is exponential in the number if contacts and may hence be infeasible to
deploy. Furthermore, its stability prediction is somewhat conservative, as we
will show later in this paper.

\section{Passive Equilibrium Formulation}

This section introduces the general framework of our problem. Consider
a grasp that consist of $m$ contacts. Each contact is defined by a
location on the surface of the object and a normal direction
(determined by the local geometry of the bodies in contact). For any
contact-specific vector (such as contact force or relative contact
motion), we will use subscript $n$ to denote the component lying in
the normal, and subscript $t$ to denote the component lying
in the tangent direction.
\begin{figure}[t]
  \centering
  \includegraphics[width=0.43\linewidth]{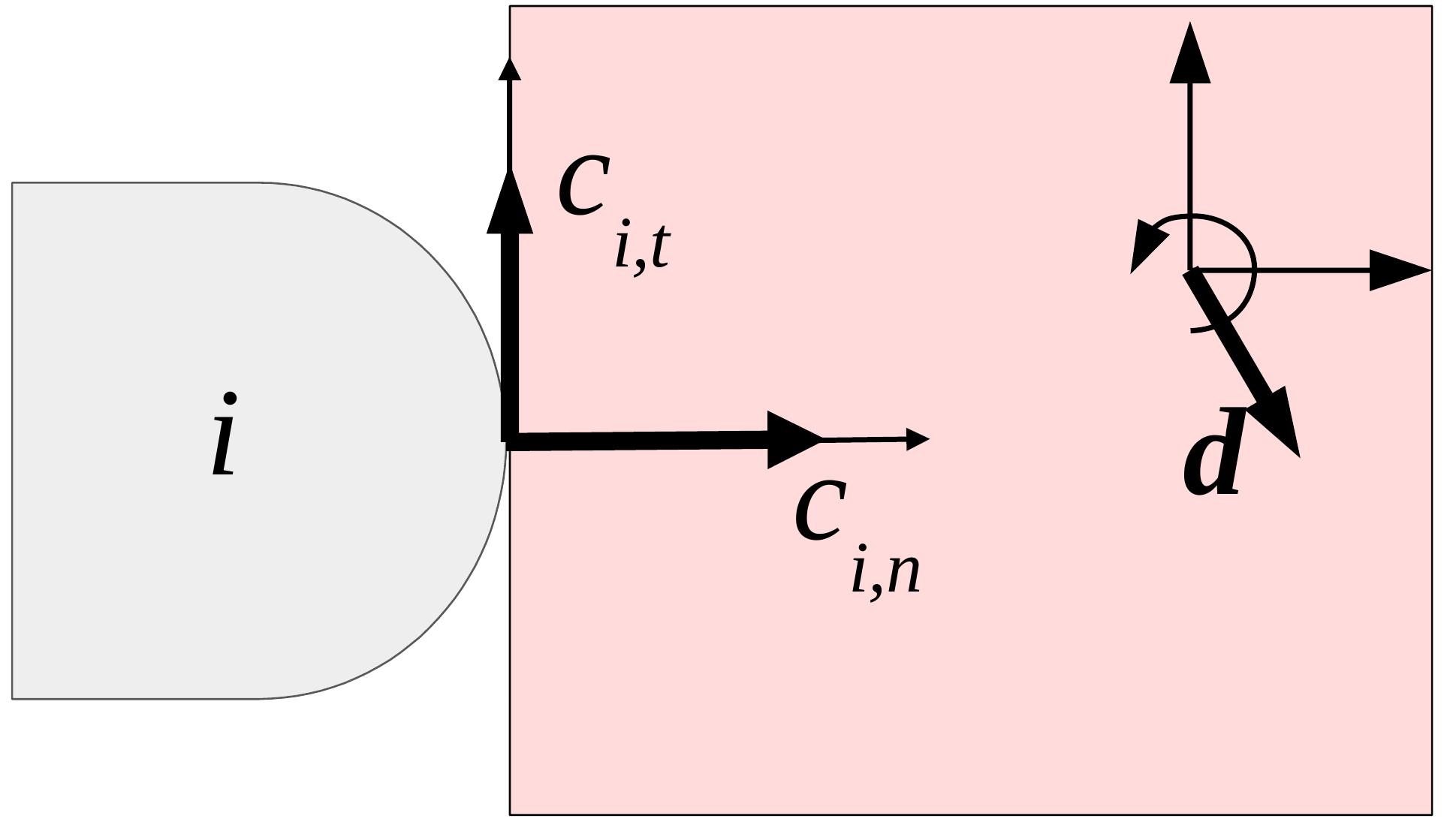} 
  \caption{Notation used the paper. At contact $i$, $c_{i,n}$ denotes
    the magnitude of the normal component of the contact force while
    $c_{i,t}$ denotes its tangential component. These are expressed in
    the contact reference frame, chosen such that one of axes aligns
    with the normal component, and the other axis with the tangential
    component. $\bm{d}$ denotes virtual object motion expressed in the
    object coordinate frame.}
  \label{fig:detail}
\end{figure}
We use the vector $\bm{c} \in \mathbb{R}^{2m}$ to denote contact
forces, where $\bm{c}_i \in \mathbb{R}^2$ is the force at the i-th
contact. Using the notation above, $c_{i,n} \in \mathbb{R}$ is the
normal component of this force, and $\bm{c}_{i,t} \in \mathbb{R}$ is
its tangential component, created via friction. We represent both as
scalars, noting that we can always choose a convenient contact
coordinate frame such that the normal direction lines up with one of
the axes. Our notation is illustrated in Fig.~\ref{fig:detail}. We now assume that the object is being disturbed by an externally
applied wrench $\bm{w} \in \mathbb{R}^3$. The first, and simplest,
condition necessary for equilibrium is the existence of a vector of
contact forces that satisfies equilibrium constraints and hence
balances the externally applied wrench. Furthermore, contact
forces have to satisfy unilaterality (only positive normal forces)
and friction constraints. At each contact $i$, the magnitude of the
friction force has to be less than or equal to the normal
force scaled by the friction coefficient $\mu$:
\begin{align}
\bm{G} \bm{c} &= \bm{w} \label{eq:equilibrium}\\
\bm{S}_n \bm{c} &\geq 0  \label{eq:unlilaterality}\\
-\mu c_{i,n} \leq c_{i,t} &\leq \mu c_{i,n} &i=1,...,m \label{eq:friction}
\end{align}
where $\bm{G} \in \mathbb{R}^{3 \times 2m}$ is the grasp map matrix
and $\bm{S}_n \in \mathbb{R}^{m \times 2m}$ is a selection matrix
which selects only the normal components from the contact wrench
vector. (In three dimensions, constraint (\ref{eq:friction}) becomes
non-linear, but is still convex.)

This problem is, in general, statically indeterminate; if a solution
exists, there are infinitely many solutions. More importantly, the existence
of a solution does not necessarily mean that the grasp is in equilibrium under
the given wrench. In the simple example of Fig.~\ref{fig:grasp} with the
upwards disturbance (shown on the left) contact forces always exist that will
satisfy (\ref{eq:equilibrium})-(\ref{eq:friction}), but, if the contacts have
not been preloaded, these forces will not arise strictly in response to the
disturbance. In other words, we are not yet capturing the passive response of
the system to the disturbance. In order to capture the passive reaction of a
grasp and resolve the indeterminacy, we start from the idea of using grasp
compliance~\cite{CUTKOSKY_COMPLIANCE}. The key idea is to add constitutive
equations that remove the indeterminacies. In particular, this is done by
introducing virtual object movement and virtual springs at the contacts.
Contact forces arise through virtual object motion loading the virtual
springs. If there are springs for each component of the contact forces, the
system becomes statically determinate: we can compute the linear stiffness of
the grasp with respect to externally applied wrenches and hence contact forces
that balance the disturbance. For now, we will neglect the compliance of the
hand mechanism and will only concern ourselves with the stiffness of the
contacts.

We begin by introducing virtual springs at the contacts, but, unlike previous
work~\cite{BICCHI93}\cite{BICCHI94}\cite{BICCHI95}\cite{PRATTICHIZZO97}, these
springs only affect forces in the normal direction. The deformation of those
springs due to a virtual object motion $\bm{d} \in \mathbb{R}^3$ determines
the normal forces at the contacts, supplying us with constitutive equations
for those forces. We introduce matrix $\bm{K}$, a stiffness
matrix for the virtual springs along the contact normals.
\begin{equation}
\bm{S}_n \bm{c} = \bm{K} \bm{S}_n \bm{G}^T \bm{d} \label{eq:constitutive}
\end{equation}
\begin{figure}[t]
  \centering
  \includegraphics[width=0.8\linewidth]{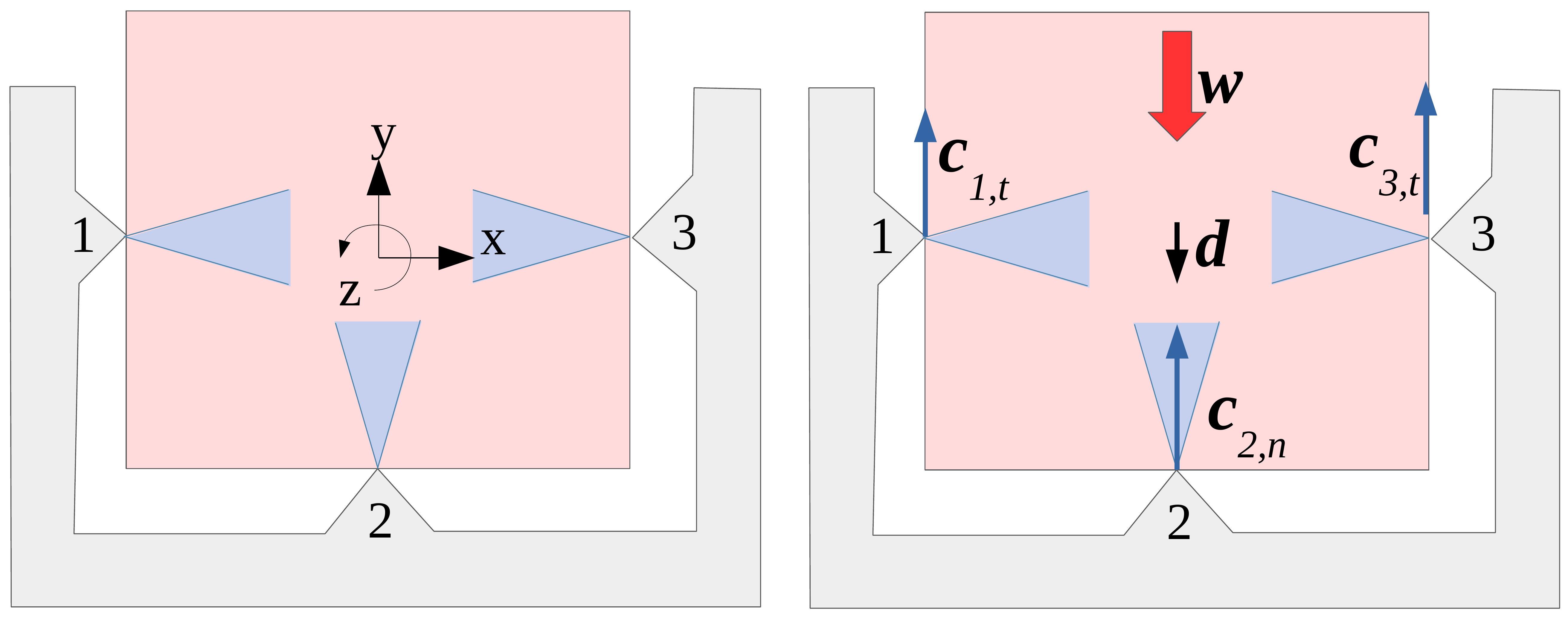} 
  \caption{Linear relationship between friction and virtual object
    motion. Left: undisturbed system with 0 contact forces. Right:
    reaction to disturbance $\bm{w}$. Downward virtual motion $\bm{d}$
    loads the normal component of $\bm{c}_2$ and resists the
    disturbance, but also creates frictional components at contacts 1
    and 3. Since $\bm{c}_1$ and $\bm{c}_3$ now violate friction
    constraints, we could erroneously conclude that the disturbance
    can not be resisted.}
  \label{fig:linear}
\end{figure}
Friction forces require additional consideration. In previous
work~\cite{BICCHI94}, these were also included in a linear
relationship such as (\ref{eq:constitutive}).  However, this approach
has a major limitation: if such a linear relationship causes friction
forces to leave the friction cone, we could erroneously conclude that
the grasp is unstable. This is the case depicted in
Fig.~\ref{fig:linear}: using a linear friction constraint would lead
us to conclude that equilibrium is not possible, where it is clear
that stable equilibrium for this combination of grasp and applied
wrench does indeed exist.\footnote{Prattichizzo et al. alleviate this
  shortcoming with an exponential complexity algorithm that allows a
  contact to slip~\cite{PRATTICHIZZO97}. However, when using this
  approach, a slipping contact may not apply any friction force at
  all, and hence their algorithm is overly conservative, as we will
  discuss later.}
\begin{figure}[t]
  \centering
  \includegraphics[width=0.8\linewidth]{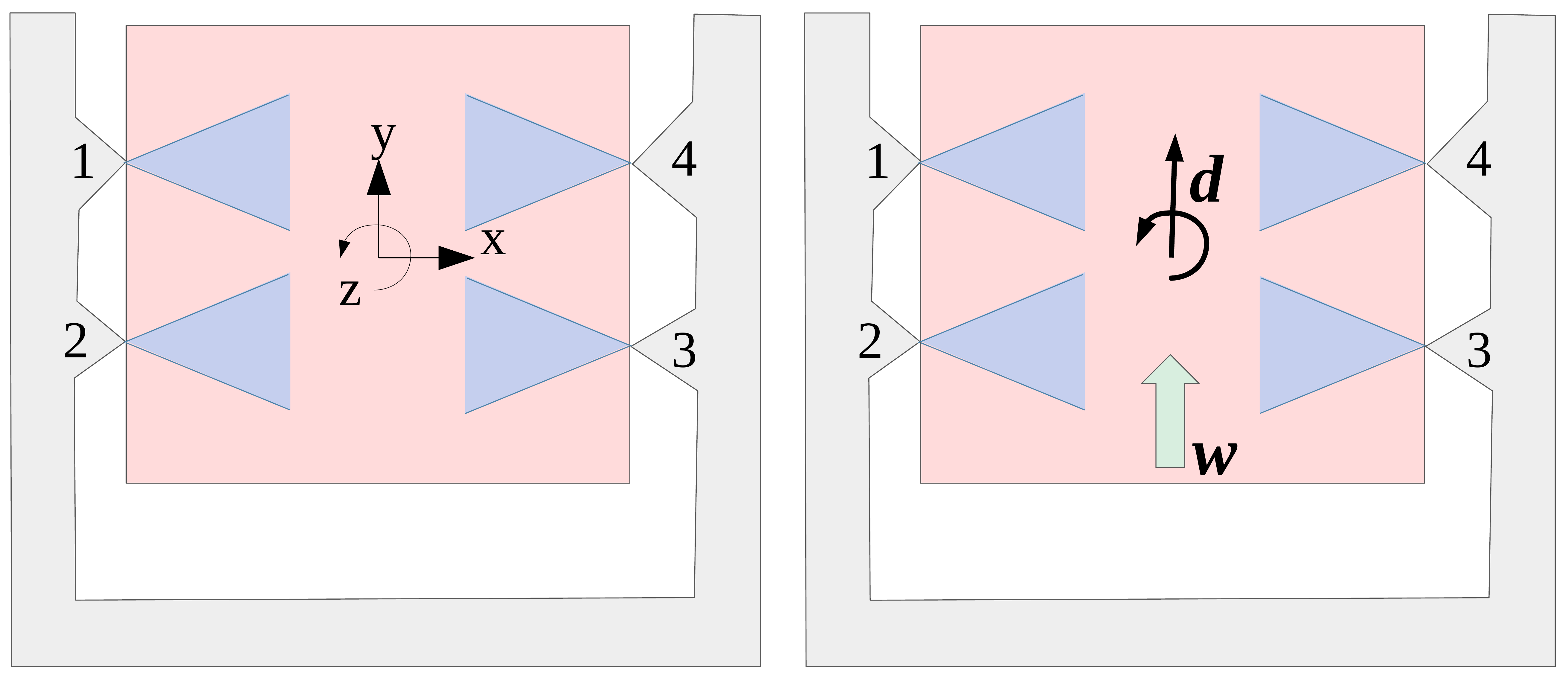} 
  \caption{Unconstrained friction forces (except for friction
    cone constraint). Left: undisturbed system with 0 contact forces (also
    showing object reference frame). Right: reaction to disturbance
    $\bm{w}$. A virtual motion that is a combination of translation
    and rotation as shown can load the contacts enough to resist an
    arbitrary wrench, even in the absence of any preloads.}
  \label{fig:twist}
\end{figure}

However, leaving friction forces constrained only by the friction cones
described by (\ref{eq:friction}) also leads to results that are physically
inconsistent. Consider the case depicted in Fig.~\ref{fig:twist}. Here, the
existence problem (are there contact forces such that constraints
(\ref{eq:equilibrium})-(\ref{eq:constitutive}) are satisfied) always has a
solution, irrespective of the magnitude of the applied wrench, or the contact
preload. The solution consists of a virtual object rotation that ``loads'' the
contacts, creating the normal forces (and thus the friction) needed to resist
the disturbance. However, in the absence of any initial preload, we would
expect system to be unstable in the presence of the shown wrench, and the
object to slide out. The underlying reason for this behavior is that we have not yet
accounted for energy constraints on our system. According to the
Maximum Dissipation Principle~\cite{PESHKIN89}, at a contact that is slipping (virtual
motion in the tangential direction is not zero) frictional force
should dissipate as much energy as possible. This is achieved if
friction opposes virtual motion, and lies on the edge of its friction
cone. Thus, at a contact $i$ that slips (and thus has relative
tangential motion), the friction force can be expressed as the vector
of relative motion multiplied by an unknown scalar $\sigma_i$, and
constrained to lie on the edge of the friction cone:
\begin{eqnarray}
  c_{i,t} &=& \sigma_i (\bm{G}^T \bm{d})_{i,t} \label{eq:sigma} \\
  \sigma_i &\leq& 0 \label{eq:sigmaneg} \\
\left|c_{i,t}\right| &=& \mu c_{i,n} \label{eq:magnitude}\\
&&\text{for all}~i~\text{s.t.}~(\bm{G}^T \bm{d})_{i,t} \neq 0 \nonumber
\end{eqnarray}
For a contact that does not slip, the friction force is still only
bound by (\ref{eq:friction}). We now also have a
constitutive relation for frictional forces, and hence all the
constraints we require to model a grasp. However, the formulation we have arrived at does not allow for an
efficient solution method: we have to distinguish between the possible
slip states of a contact in order to decide which constraints
apply. With $m$ contacts, considering only two (stick/slip) possible
states for each contact still leaves us with a total of $2^m$ possible
combinations for our system.
    
In this light, our contributions are as follows. For two-dimensional grasps,
we will show that, given rigid body movement constraints, the \textit{number
of possible slip states for the system is in fact   polynomial in the number
of contacts}, even if also including the direction (positive or negative) of
slip along a tangential axis. Then, we will show how this result allows us to
\textit{break the problem down into a polynomial number of sub-problems}, which can be solved efficiently.

\section{Number of Possible Slip States}
\begin{figure*}[t!]
  \centering
  \includegraphics[width=1.0\linewidth]{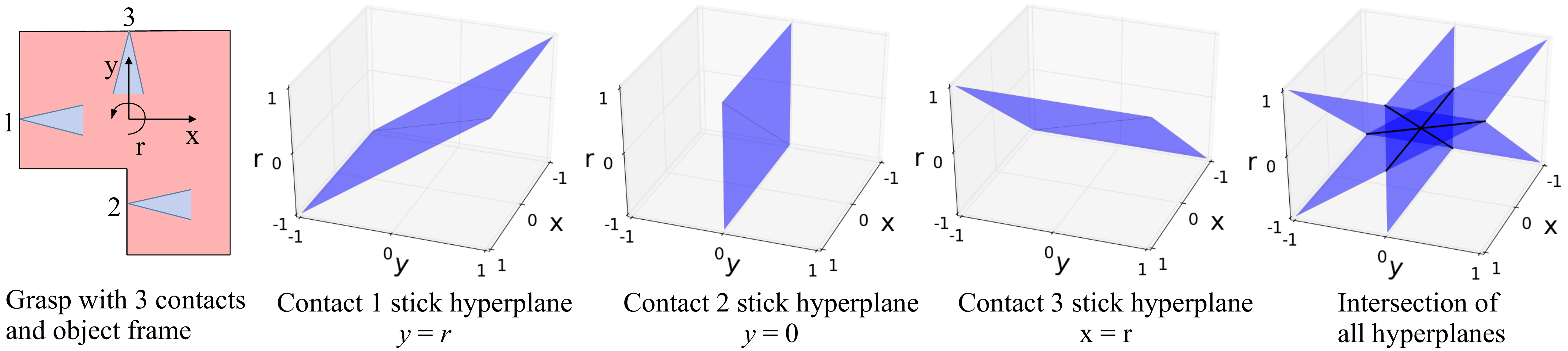}
  \vspace{-5mm}
  \caption{Arrangement of planes equivalent to contact stick-slip
    constraints. For the grasp shown on the left, each contact defines
    a plane in the three-dimensional space of possible object motions
    $\bm{d}=[x,y,r]$. For example, Contact 1 sticks if the translation
    component of $\bm{d}$ along $y$ is counteracted by an equal
    rotational component $r$. Similarly, Contact 2 sticks if the $y$
    component of $\bm{d}$ is 0, and so on.}
  \label{fig:planes}
\end{figure*}

We focus here on the problem of determining the state of each contact
for a two-dimensional grasp with $m$ contacts. In its simplest form,
this state only considers two possibilities per contact: sticking or
slipping. However, we consider three possible states for
each contact: stick, plus slip in the positive or negative direction
of the local tangent axis. We define a \textbf{slip state} $S_k \in \{-1,0,1\}^m$ as a vector
comprising information about slip at every contact. The $i$-th element
of $S_k$, labeled $s_i^k$, defines the state of contact $i$ in state
$k$ as follows:
\begin{eqnarray}
  s_i^k = 0&:& \text{stick}, (\bm{G}^T \bm{d})_{i,t} = 0 \\
  s_i^k = -1&:& \text{negative slip}, (\bm{G}^T \bm{d})_{i,t} < 0 \\
  s_i^k = 1&:& \text{positive slip}, (\bm{G}^T \bm{d})_{i,t} > 0
\end{eqnarray}
Finally $\mathbb{S}$ is the set of all possible system slip states,
thus $S_k \in \mathbb{S}$ for $k=1..\#(\mathbb{S})$, where
$\#(\mathbb{S})$ is the cardinality of $\mathbb{S}$. At first glance, $\#(\mathbb{S})=3^m$: since each contact can have
three states, the total number of states for the system is exponential
in the number of contacts. Indeed, this is the approach used in
previous studies that account for stick/slip at each
contact~\cite{PRATTICHIZZO97}. Our main insight is that not all of these possible state combinations
are consistent with rigid body movement of the grasped
object. Assuming that the object is rigid, displacement at each
contact reference frame must be related to object displacement at all
other contacts, and a linear function of object displacement $\bm{d}$
expressed at the object reference frame. In fact, we will show that,
when accounting for rigid body motion, $\#(\mathbb{S})$ is quadratic
in $m$.

\subsection{Slip states as plane arrangements}

We make an argument from geometry to show that not all combinations of
slip states are indeed possible. First, let us look at the stick
condition defined above: in the three-dimensional space of possible
object motions $\bm{d}=[x,y,r]$, the constraint $(\bm{G}^T
\bm{d})_{i,t} = 0$ defines a plane. Note that this
plane goes through the origin. Any object motion lying on this plane
will result in zero relative tangential motion at this contact. Motion
in the halfspace where $(\bm{G}^T \bm{d})_{i,t} \geq 0$ will result in
slip along the tangential axis in the positive direction, while motion in
the complementary halfspace $(\bm{G}^T \bm{d})_{i,t} \leq 0$ will
result in slip in the negative direction. Combining the planes defined by each contact, we obtain
the possible states for all of our contacts. These planes
segment the space of object motions into:
\begin{itemize}
  \item 3-dimensional ``regions'' where all contacts are slipping;
  \item 2-dimensional ``facets'' (region boundaries on a single
  plane) where one contact is sticking;
  \item 1-dimensional ``lines'' (intersections of multiple
    planes) where multiple contacts are sticking.
\end{itemize}
By construction, since all of our planes go through the origin,
the only possible zero-dimensional ``point'' intersection is the
origin itself (see Fig.~\ref{fig:planes}.)
Given this partition of the space of possible object motions, it
follows that any system slip state $S_k$ that is consistent with a
possible object motion must correspond to either a region, a face, or
a line created by this plane arrangement. Finding the maximum number of three-dimensional regions given $m$
planes is equivalent to finding the maximum number of
regions on a sphere cut with $m$ great circles, which is known to be
$O(m^2)$~\cite{GREATCIRCLES}. However, the regions do not define all the combinations of
slip states we care about. We must also consider the cases where at
least one contact sticks, namely the ``facets'' and ``lines'' defined
as above. 
\begin{figure}[t]
  \centering
  \includegraphics[width=0.9\linewidth]{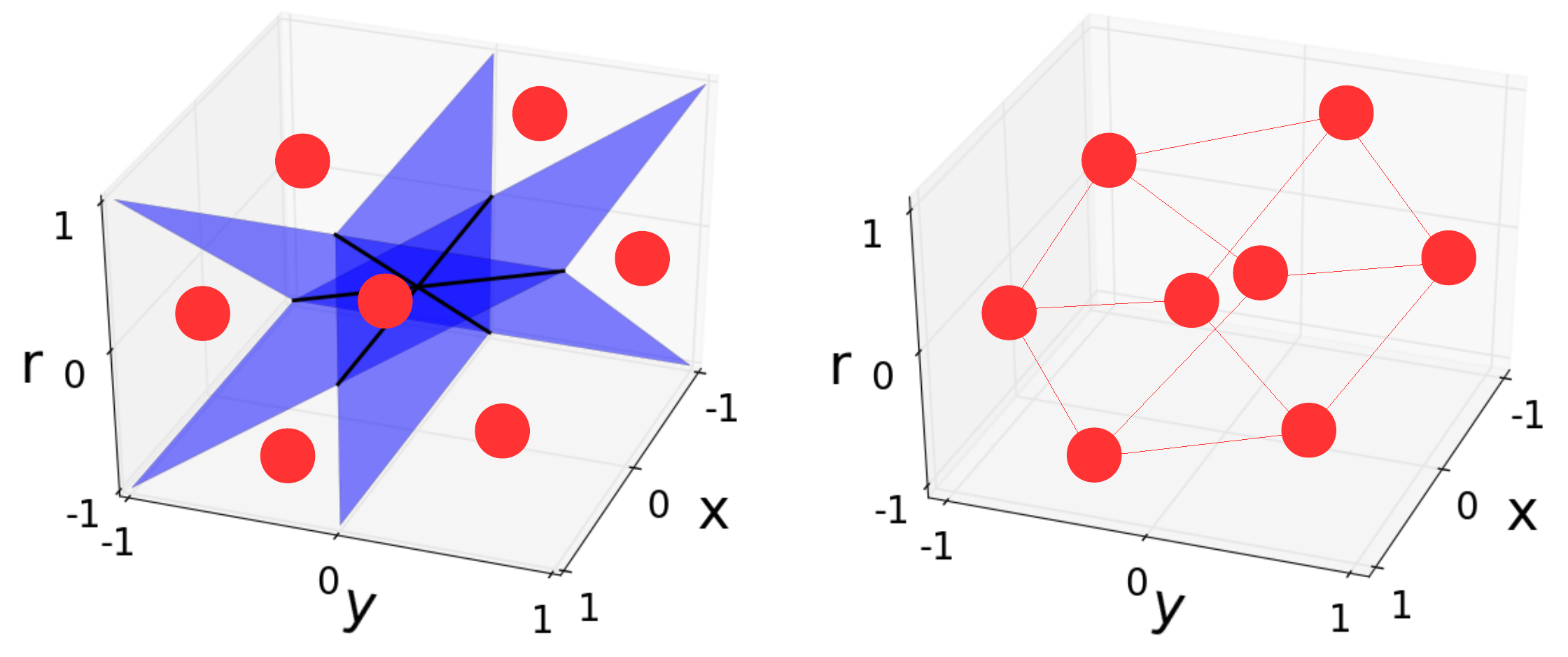} 
  \caption{Dual polyhedron for the arrangement of planes. Each region
    of the arrangement of planes (red dot) corresponds to a
    vertex of the polyhedron (for clarity, only 7 regions are marked
    in the left figure). Vertices are connected if the corresponding
    regions share a facet.}
  \label{fig:poly}
\end{figure}
We can show that the number of regions, facets and lines is
bounded polynomially by applying Zaslavski's formula~\cite{FUKUDA91}. Let
$f_k^{(d)}(n)$ be the number of ``k-faces'' of an arrangement of $n$ hyperplanes
in $d$ dimensional space, where $k$ is the dimension of the face. Then, the following holds:
\begin{equation}
f_k^{(d)}(n) \leq {n \choose d-k} \sum_{i=0}^k {n-d+k \choose i}
\end{equation}
In our case, the total number of slip states we are interested in is
equal to $\sum_{k=0}^3 f_k^{(3)}(m)$ and hence polynomial in $m$. This is, however, an upper bound that will not be
attained in our case, as we show next. Let us construct the dual convex polyhedron to our arrangement of
planes, which will be instrumental in enumerating all possible slip
states. Each region of the plane arrangement corresponds to a vertex
of the polyhedron, and each two dimensional boundary (facet) between
regions corresponds to an edge (vertices corresponding to neighboring
regions are connected, see Fig.~\ref{fig:poly}.) We can ensure this polyhedron is convex by
selecting, as the representative vertex for each region, a point where
the region intersects the unit sphere. We note that the lines of
our plane arrangement correspond to faces of the dual polyhedron. The
dual polyhedron thus fully describes our possible slip states. 

Like any convex polyhedron, the dual we have constructed can be represented by a
3-connected planar graph. From this result, we can bound number of slip states even more
closely, than with Zaslavski's formula: Any maximal planar graph with $V$
vertices has at most $3V-6$ edges and $2V-4$ faces and hence the number of
edges and faces of our polyhedron are linearly bounded by the number its
vertices. Since we have shown the number of vertices to be $O(m^2)$, so are
the number of edges and faces. Thus, the number of slip states we must consider
is quadratic in the number of contacts.

\subsection{Slip state enumeration}

We can now present a complete procedure for enumerating all possible
slip states $S_k$ of an $m$-contact system that are consistent with
rigid body object motion.

\mystep{Step 1.} We begin by enumerating all the slip states $S_k$
corresponding to regions in our plane arrangement. We achieve that
using Algorithm~\ref{alg:buildset}. We note that for any state $S_k$
obtained by this algorithm, all contacts are slipping, in either the
positive or negative direction ($s_i^k=\pm 1$ for all $i$, we have not
yet considered sticking cases).
\begin{algorithm}[!t]
\caption{}\label{alg:buildset}
\begin{algorithmic}[0]
\State Initialize $\mathbb{S}$ with empty state $S$
\For{$i=1..m$}
  \For{$k=1..\#(S)$}
    \State Remove $S_k$ from $\mathbb{S}$
    \If{Region $S_k$ has component above plane $P_i$}
      \State Create region $S_{k+} = \{S_k, 1\}$
      \State Add $S_{k+}$ to $\mathbb{S}$
    \EndIf
    \If{Region $S_k$ has component below plane $P_i$}
      \State Create region $S_{k-} = \{S_k, -1\}$
      \State Add $S_{k-}$ to $\mathbb{S}$
    \EndIf
  \EndFor
\EndFor  
\end{algorithmic}
\end{algorithm}

\mystep{Step 2.} Now we create the slip states corresponding to facets
in the plane arrangements (one sticking contact). As mentioned before,
these correspond to edges of the dual polyhedron, so we begin this
step by constructing the dual polyhedron. We already have its
vertices: each state $S$ created at the previous step defines a region
of the plane arrangement, and thus corresponds to a vertex of the
dual polyhedron. Then, for every two states $S_k, S_l$ in
$\mathbb{S}$ that differ by a single $s_i$, we add the edge between
them to the dual polyhedron (note that thus our polyhedron is also a
partial cube where edges connect any two vertices with Hamming
distance equal to 1~\cite{EPPSTEIN08}). Furthermore, we also create an
additional state $S_{kl}$ corresponding to the facet between $S_k$ and
$S_l$. This will be identical to both $S_k$ and $S_l$, with the
difference that $s_i=0$ (the entry corresponding to the plane that
this facet is on is set to 0).

\mystep{Step 3.} Finally, we must create the slip states corresponding to
lines in our plane arrangement (multiple sticking contacts), which correspond
to faces of the dual polyhedron. We obtain the faces of our dual polyhedron by
computing the Minimum Cycle Basis (MCB) of the undirected graph defined by its
edges (computed at the previous step). This gives us $F-1$ of the faces of our
polyhedron; to see why consider that the number of cycles in the minimum cycle
basis is given by $E-V+1$~\cite{MEHLHORN05}. Recall the Euler-Poincar\'e
characteristic $\chi = V - E + F$ relating the number of vertices, edges and
faces of a manifold. For a convex polyhedron $\chi = 2$, and from this we can
derive the number of faces of our dual polyhedron to be equal to $E-V+2$. The
last face is obtained as the symmetric sum of all the cycles in the MCB
(defined as in~\cite{MEHLHORN05}).

Once we have the cycles corresponding to the faces of the dual
polyhedron, we convert them into slip states as follows. For each
cycle, starting from the slip state $S_k$ corresponding to any of the
vertices in the cycle, we set $s_i=0$ for any plane $i$ that is
traversed by an edge in the cycle. The total number of slip states $S$ we obtain is thus equal to the
number of regions, facets and lines of the plane arrangement, which is
the same as the number of vertices, edges and faces of its dual
polyhedron. We have already shown that this is polynomial (quadratic) in the
number of planes (contacts). We also show that the enumeration
algorithm above has polynomial runtime.

We note that Step 1 has two nested loops, with one iterating over
planes and the other one over existing states. The number of states at
the end of this step is bounded by $m^2$, thus the running time of
this Step is $O(m^3)$. Step 2 must check every state against every
other one, with $O(m^2)$ states, thus its running time is
$O(m^4)$. Finally, the dominant part of Step 3 is the computation of
the MCB. We have used an implementation with $O(E^3 + VE^2\text{log}
V)$ running time, where $V$ and $E$ are the number of vertices and
edges of the dual polyhedron. Since both $E$ and $V$ are polynomial in
$m$, the running time of the MCB algorithm is as well. Thus our
complete enumeration method has polynomial runtime in the number of
contacts $m$.

\begin{algorithm}[!t]
\caption{}\label{alg:complete}
\begin{algorithmic}[0]
\State Build $\mathbb{S}$, the set of all possible slip states
\For{$k=1..\#(S)$}
\State Given $S_k$, solve system (\ref{eq:equilibrium}), (\ref{eq:unlilaterality}), (\ref{eq:constitutive}), (\ref{eq:plus}-\ref{eq:zero})
\If{solution found}
\State \textbf{Return:} grasp stable
\EndIf
\EndFor
\State \textbf{Return:} grasp unstable
\end{algorithmic}
\end{algorithm}

\section{Complete Equilibrium Determination}

In this section we use the set of slip states $\mathbb{S}$ previously derived to arrive at a complete algorithm for determining the existence
of passive equilibrium for our system.

\subsection{Solution for a particular slip state}

We recall that a slip state $S_k=\{s_i\}, i \in \{1..m\}$, where
$s_i^k=\{-1,0,1\}$. For each contact $i$, $s_i^k=-1$ means that the
contact is slipping in the negative direction, $s_i^k=1$ means slip in
the positive direction, and finally $s_i^k=0$ means the contact is
sticking. Critically, the fact that $S_k$ comprises not just stick/slip
information, but also the direction of slip, turns friction into a
simple linear dependency on object motion. The complete friction
constraints are:
\begin{eqnarray}
  c_{i,t} = \mu c_{i,n}&~~~\text{for all}~i~\text{s.t.}~s_i^k=-1 \label{eq:plus}\\
  c_{i,t} = -\mu c_{i,n}&~~~\text{for all}~i~\text{s.t.}~s_i^k=1 \label{eq:minus}\\
  -\mu c_{i,n} \leq c_{i,t} \leq \mu c_{i,n}&~~~\text{for all}~i~\text{s.t.}~s_i^k=0 \label{eq:zero}
\end{eqnarray}
Intuitively, these correspond to the following three states:
\begin{enumerate}
\item The contact slips in the negative tangential direction. Friction opposes the relative motion s.t. $c_{i,t} = \mu
  c_{i,n}$.
\item The contact slips in the positive tangential direction. Friction opposes the relative motion s.t. $c_{i,t} = -\mu c_{i,n}$.
\item The contact sticks (it exhibits no relative motion in the
  tangential direction). The friction force lies in the
  interior of the friction cone s.t. $-\mu c_{j,n} \leq c_{j,t}
  \leq \mu c_{jn}$.
\end{enumerate}
Thus, for any given slip state $S_k$, the system given by
Eqs. (\ref{eq:equilibrium}), (\ref{eq:unlilaterality}),
(\ref{eq:constitutive}) and (\ref{eq:plus}-\ref{eq:zero}) is easy to
solve. In fact, we notice that we have a total of $3 + 2m$ unknowns
(vectors $\bm{d}$ and $\bm{c}$). From equilibrium in
eq. (\ref{eq:equilibrium}), the normal force constitutive relations in
eq. (\ref{eq:constitutive}) and the above
eq. (\ref{eq:plus}\&\ref{eq:minus}) we have $3 + m + (m-l)$
constraints, where $l$ is the number of slipping contacts. We can use
the condition that the remaining $m - l$ stationary contacts do not
exhibit any relative motion in the tangential direction to formulate
$m - l$ additional constraints, leading to a total of $3 + 2m$
constraints matching the number of unknowns. For a given $S_k$ we can thus trivially find a solution to the linear system
of equations given by Eqs. (\ref{eq:equilibrium}),(\ref{eq:constitutive}) and
(\ref{eq:plus}\&\ref{eq:minus}). We then check if the solution also meets Eqs.
(\ref{eq:zero}) (friction cone at sticking contacts) and
(\ref{eq:unlilaterality}) (normal force non-negativity). If the matrix
corresponding to Eqs. (\ref{eq:equilibrium}),(\ref{eq:constitutive}) and
(\ref{eq:plus}\&\ref{eq:minus}) is singular this operation is replaced by a
simple linear program, which looks for a solution in the nullspace satisfying
also equations (\ref{eq:zero}),(\ref{eq:unlilaterality}). Only if the solution we obtain satisfies all constraints do we deem the grasp to
be stable for slip state $S_k$.

\subsection{Complete equilibrium algorithm}

We can now formalize our complete algorithm using the components
outlined so far (Algorithm~\ref{alg:complete}). We first build the total set of possible slip states
$\mathbb{S}$. Then, for every $S_k \in \mathbb{S}$, we check for a
solution to the system described above. If one exists, we deem the
grasp stable. If, after enumerating all possible $S_k$, we do not find
one that admits a solution, we deem the grasp unstable. We make two important observations regarding
Algorithm~\ref{alg:complete}. First, its running time is polynomial in
the number $m$ of contacts. This follows trivially from the results
obtained so far. We know that $\#(\mathbb{S})$ is polynomial
in $m$, as is the process for building it. For each $S_k$, we then
solve at most a linear program with $2m+3$ unknowns, which also has
a polynomial runtime, which completes this result.

Second, Algorithm~\ref{alg:complete} guarantees that, if no solution
is found, none exists that satisfies the constraints of our
system. $\mathbb{S}$ provably contains all the slip states consistent
with rigid body movement; for each of these, equilibrium conditions
form a linear program for which we can provably find all solutions (if
they exist). So, under the assumed formulation (virtual springs used
to determine passive reaction, and frictional constraints including
the maximum dissipation principle), if a solution exists to the
equilibrium problem, we must find it.


\begin{figure*}[t!]
\centering
\subfigure[Slice through the three dimensional GWS for zero applied torque]{\label{fig:GWS}\includegraphics[width=0.32\linewidth]{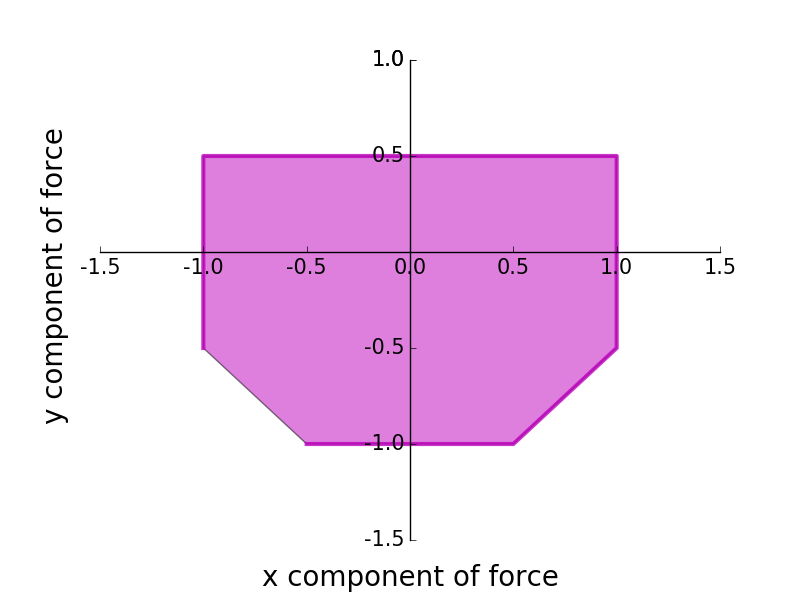}}
\subfigure[Resistible forces with no preload (our algorithm)]{\label{fig:bin_no_preload}\includegraphics[width=0.32\linewidth]{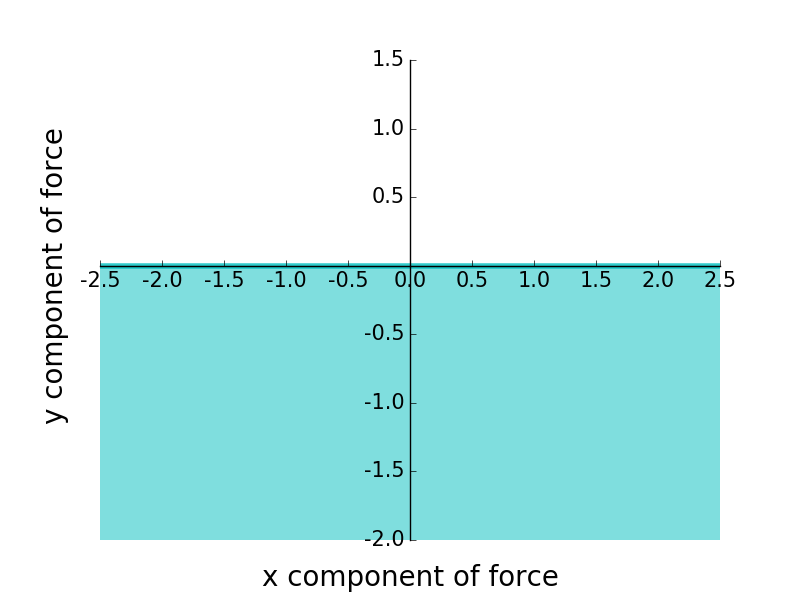}}
\subfigure[Resistible forces with a preload such that the normal force at each conctact is 1 (our algorithm)]{\label{fig:bin_preload}\includegraphics[width=0.32\linewidth]{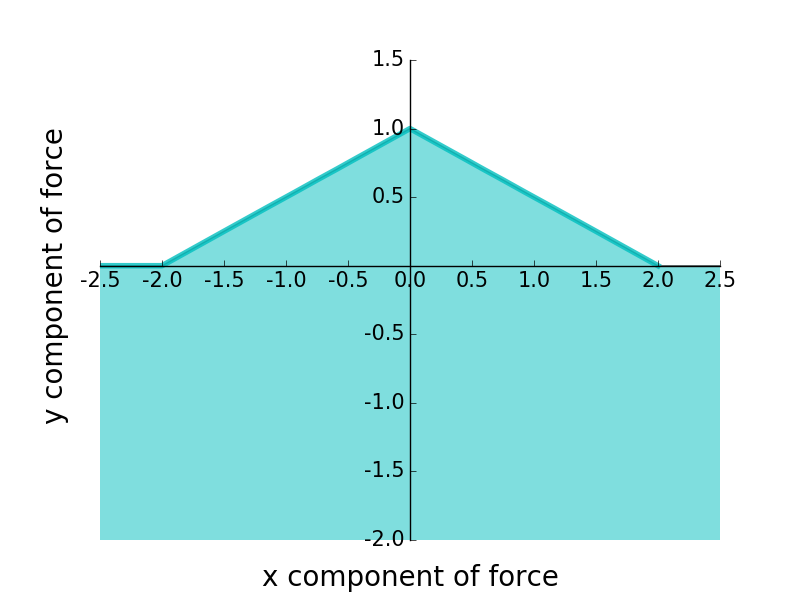}}
\caption{Grasp stability representations for the grasp in Fig.
  \ref{fig:grasp}. The GWS representation (a) shows the space of
  applied wrenches the grasp can resist with contact forces that
  satisfy only friction constraints (shaded). We are using the L1 norm GWS,
  meaning the normal component of a contact force is at most equal to
  1. Our algorithm (b)(c) captures passive resistance up to arbitrary
  reaction forces, as long as they arise passively in response to the
  disturbance. The space of resistible wrenches produced by our
  algorithm is thus bounded only on the upper edge of the space, as
  indicated by the line border. A boundary without this line indicates
  arbitrary resistance in the corresponding directions.}
\end{figure*}

\section{RESULTS}

In this section we will demonstrate that our framework - in contrast to
previous approaches to this problem - predicts the correct force distributions
and makes an accurate prediction on grasp stability. We will utilize the grasp
examples introduced in Figs. \ref{fig:grasp}, \ref{fig:linear} \&
\ref{fig:twist}, because the correct force distribution and stability of the
grasp is easily understood intuitively.

Let us first consider the problem first described in the Introduction: can we
discriminate which applied wrenches will be balanced purely passively, and
where an active preload of the grasp is required? Recall that there exist
contact forces in the interior of the friction cones that balance both
wrenches shown in Fig. \ref{fig:grasp}. Perhaps the most commonly used
approach to grasp stability analysis is the Grasp Wrench Space
method~\cite{FERRARI92}. Indeed, when we consider the slice through the GWS
visualized in Fig. \ref{fig:GWS} we can see that there exist contact forces
that balance arbitrary forces in the plane. However, we argue that while
$\bm{w}_2$ will always be reacted passively, no matter the preload, in order
to react $\bm{w}_1$ we require the grasp to have been sufficiently preloaded.
The GWS method correctly indicates the existence of equilibrium contact forces
but does not predict if they may arise, and hence does not capture the
necessity of a preload.

Now let us apply our framework to this problem: Using our algorithm, we can
test the resistance of this grasp to forces in the plane. We do this by
discretizing the direction of application of force to the object and finding the maximum
resistible force in each direction using a binary search. Figs.
\ref{fig:bin_no_preload} \& \ref{fig:bin_preload} shows the region of
resistible wrenches for a grasp without and with a preload respectively. As
our algorithm takes into account passive effects it correctly predicts that, in
both cases, forces with non-positive component in the y-axis and arbitrary
magnitude can be withstood. Indeed, for any applied wrench $\bm{w}=(0,w_y,0),
w_y \leq 0$ our framework predicts contact forces $(0,0)$ at contacts 1 and 3,
and contact force $(-w_y,0)$ at contact 2 (see Table \ref{tab:three_contact}).
Furthermore, it captures the necessity of a preload in order to resist forces
with positive component in the y-direction: For $w_y > 0$ our algorithm finds
no solution, and hence the grasp must be unstable to this disturbance, unless
an appropriate preload is applied.

We have already shown the main deficiency of compliance based
approaches such as
in~\cite{BICCHI93}\cite{BICCHI94}\cite{BICCHI95}\cite{PRATTICHIZZO97}
(recall Fig. \ref{fig:linear}), which are commonly used to predict
contact forces that arise in grasps due to disturbances. We can now
use this grasp (Fig. \ref{fig:grasp}) and our algorithm to show that
the improvements to a linear compliance suggested by Prattichizzo et
al.~\cite{PRATTICHIZZO97} lead to overly conservative stability
estimates.  Their approach allows each contact to be in one of
three states: sticking, slipping or detached. A slipping contact may
not apply any frictional forces, while a detached contact may not
apply any force at all. If we try every possible combination of states
and modify the compliance of the grasp accordingly, this alleviates
some of the problems of the purely linear compliance approach: If we
consider contacts 1 \& 3 to be slipping, our grasp in Fig.
\ref{fig:grasp} may now withstand arbitrary forces, where $w_y \leq
0$.

This approach, however, does not allow us to arrive at the correct
result in cases where $w_y \geq 0$. Consider the preloaded grasp
before the application of an external wrench (Table
\ref{tab:three_contact}.) The contact forces on both contacts 1 \& 3
lie on the friction cone edge in order to balance the preload applied
by contact 2. If we now apply an external wrench $\bm{w}=(0,1,0)$,
there exists no combination of sticking, slipping and detached
contacts (and corresponding modifications of the linear compliance)
that results in legal contact forces. Our algorithm, however, predicts
a stable grasp (Table \ref{tab:three_contact}), showing how important
friction is for grasp stability and why a correct treatment of
friction is fundamental to stability analysis. Furthermore, we have
arrived at this result in polynomial time --- we did not have to
consider exponentially many slip states, as in~\cite{PRATTICHIZZO97}.
\begin{table}[t]
\centering
\begin{tabular}{l|l|l|lll|l}
$\bm{w}$ & $P$ & Stable & $\bm{f}_1$ & $\bm{f}_2$ & $\bm{f}_3$ & $\bm{d}$ \\
\hline
$(0, 0, 0)$ & 0 & Yes & $(0,0)$ & $(0,0)$ & $(0,0)$ & $(0,0,0)$ \\
$(0, 0, 0)$ & 1 & Yes & $(1,-0.5)$ & $(1,0)$ & $(1,0.5)$ & $(0,0,0)$ \\
$(0, -1, 0)$ & 0 & Yes & $(0,0)$ & $(1,0)$ & $(0,0)$ & $(0,-1,0)$ \\
$(0, -2, 0)$ & 0 & Yes & $(0,0)$ & $(2,0)$ & $(0,0)$ & $(0,-2,0)$ \\
$(0, 1, 0)$ & 0 & No & - & - & - & - \\
$(0, 1, 0)$ & 1 & Yes & $(1,-0.5)$ & $(0,0)$ & $(1,0.5)$ & $(0,1,0)$ \\
$(0, 1.1, 0)$ & 1 & No & - & - & - & - \\
\end{tabular}
\caption{Contact forces $\bm{f}_i=(f_{n,i},f_{t,i})$ and virtual object motion $\bm{d}=(x,y,r)$ for the grasp in Fig. \ref{fig:grasp} and a range of applied wrenches $\bm{w}=(w_x,w_y,w_z)$. The preload $P$ is such that the normal force at each contact is equal to either 0 or 1 before any wrench is applied. The object motion and applied wrenches are expressed in the coordinate frame shown in Fig. \ref{fig:grasp} and contact forces are expressed in frames as shown in Fig. \ref{fig:detail}.}
\label{tab:three_contact}
\end{table}
\begin{table}[t]
\centering
\begin{tabular}{l|lllllllll}
$m$              & 2 & 3 & 4 & 5 & 6 & 7 & 8 & 9 \\
\hline
$\#(\mathbb{S})$ & 10 & 26 & 50 & 82 & 122 & 170 & 226 & 290 \\
\hline
$time (s)$       & 0.006 & 0.02 & 0.09 & 0.42 & 1.5 & 4.8 & 13.8 & 35.7 \\
\end{tabular}
\caption{Number of slip states $\#(\mathbb{S})$ and computation time for grasps with $m$ randomly generated contacts.}
\label{tab:performance}
\end{table}
\begin{table*}[t]
\centering
\begin{tabular}{l|l|l|llll|l}
$\bm{w}$ & $P$ & Stable & $\bm{f}_1$ & $\bm{f}_2$ & $\bm{f}_3$ & $\bm{f}_4$ & $\bm{d}$ \\
\hline
$(0, 0, 0)$ & $0$ & Yes & $(0,0)$ & $(0,0)$ & $(0,0)$ & $(0,0)$ & $(0,0,0)$ \\
$(0, 0, 0)$ & $1$ & Yes & $(1,0)$ & $(1,0)$ & $(1,0)$ & $(1,0)$ & $(0,0,0)$ \\
$(0, 2, 0)$ & $0$ & No & - & - & - & - & - \\
$(0, -2, 0)$ & $0$ & No & - & - & - & - & - \\
$(0, 2, 0)$ & $1$ & Yes & $(1,-0.5)$ & $(1,-0.5)$ & $(1,0.5)$ & $(1,0.5)$ & $(0,0,0)$ \\
$(0, -2, 0)$ & $1$ & Yes & $(1,0.5)$ & $(1,0.5)$ & $(1,-0.5)$ & $(1,-0.5)$ & $(0,0,0)$ \\
$(0, 0,3)$ & $0$ & Yes & $(1,0.5)$ & $(0,0)$ & $(1,0.5)$ & $(0,0)$ & $(0,-1,1)$ \\
$(0, 0,-3)$ & $0$ & Yes & $(0,0)$ & $(1,-0.5)$ & $(0,0)$ & $(1,-0.5)$ & $(0,1,-1)$ \\
$(0, 0,3)$ & $1$ & Yes & $(1.25,0.625)$ & $(0.75,0.375)$ & $(1.25,0.625)$ & $(0.75,0.375)$ & $(0,-0.25,0.25)$ \\
$(0, 0,-3)$ & $1$ & Yes & $(0.75,-0.375)$ & $(1.25,-0.625)$ & $(0.75,-0.375)$ & $(1.25,-0.625)$ & $(0,0.0.25,-0.25)$ \\
\end{tabular}
\caption{Contact forces $\bm{f}_i=(f_{n,i},f_{t,i})$ and virtual object motion $\bm{d}=(x,y,r)$ for the grasp in Fig. \ref{fig:twist} and a range of applied wrenches $\bm{w}=(w_x,w_y,w_z)$ and preloads $P$. The object motion and applied wrenches are expressed in the coordinate frame shown in Fig. \ref{fig:twist} and contact forces are expressed in frames as shown in Fig. \ref{fig:detail}.}
\label{tab:four_contact}
\end{table*}

Let us now look at the grasp in Fig. \ref{fig:twist},
with which we investigated the deficiency of solely constraining the friction
forces to lie within the friction cones. These constraints would allow a
solver to rotate the object; the normal forces built up this way allow for arbitrary
friction forces and reaction of arbitrary wrenches. In contrast, our framework restricts the friction force as is physically correct, and negates
equilibrium for any wrench $\bm{w}=(0,w_y,0)$ if there is no preload (see Table \ref{tab:four_contact}). Hence our algorithm correctly predicts
that the contact forces required to balance this wrench will not arise
passively.

Let us now apply a preload to the contacts such that the normal force at each
contact is 1 (previous to the application of an external wrench). If we now
apply $(0,w_y,0)$ our algorithm predicts stability for $\|w_y\| \leq 4\mu$
only, where $\mu$ is the friction coefficient of the contacts (we have chosen
$\mu=0.5$). Table \ref{tab:four_contact} contains a summary of contact forces
and virtual object motion for a range of different applied wrenches and
preloads.

From a computational effort perspective, a summary of the performance
of our algorithm for enumeration of slip states can be found in Table
\ref{tab:performance}. All computation was performed on a commodity
computer with a 2.80GHz Inter Core i7 processor.

\section{CONCLUSIONS}

We have discussed the various pitfalls of physically accurate force
distribution and grasp stability analysis that also provides strong
guarantees. In order to solve this problem we considered contact slip states
(where each contact is labeled as sticking or slipping, and the direction of
slip is also determined) that are consistent with rigid body motion. The
number of possible combinations of such states was shown to be quadratic in
the number of contacts. Furthermore, we described a polynomial runtime
algorithm to enumerate those slip states. We then showed the problem of force distribution for a two dimensional
grasp to be efficiently solvable given the slip states of its contacts. Thus we
broke the force distribution problem into a number of sub-problems,
which we can efficiently solve. Recalling the above result that the number
of these problems to be solved is quadratic in the number of contacts, we used
this insight to develop an algorithm to investigate the distinction between
active and passive reactions in grasping. This is a powerful tool for
stability analysis, which in previous work required either approximation or the loss of global guarantees with respect to stability - our approach requires neither.

It is important to note that the problem of force distribution in
general 3-dimensional space is much harder. Due to the non-convexity of the
friction constraint derived from the maximum dissipation principle one cannot
efficiently solve the problem \textit{even with prior knowledge of which
contacts slip and which do not}. While in two dimensions, two tangential axes
will positively span the space of relative tangential motion (and there
are hence 3 slip states per contact), in three dimensions there are infinitely
many directions a contact can slip in. The enumeration of slip states in
3D grasping is nonetheless highly interesting and future work will
focus on generalizing the algorithm presented in this paper to higher
dimensions. Defining a high number of hyperplanes per contact could be a
powerful tool in developing efficient algorithms that compute approximate
solutions to the 3-dimensional grasping problem. In addition, we believe that our method of accounting for slip patterns
through the geometry of hyperplane arrangements may be useful in improving the performance of quality metrics such as PCR and PGR, but also elsewhere
in robotics.

\section*{Acknowledgment}

This work was supported in part by ONR Young Investigator Program award N00014-16-1-2026 and NSF CAREER Award 1551631.

\bibliographystyle{plainnat}
\bibliography{bib/orthosis,bib/grasping,bib/thesis}

\begin{thebibliography}{15}
\providecommand{\natexlab}[1]{#1}
\providecommand{\url}[1]{\texttt{#1}}
\expandafter\ifx\csname urlstyle\endcsname\relax
  \providecommand{\doi}[1]{doi: #1}\else
  \providecommand{\doi}{doi: \begingroup \urlstyle{rm}\Url}\fi

\bibitem[Bicchi(1993)]{BICCHI93}
Antonio Bicchi.
\newblock Force distribution in multiple whole-limb manipulation.
\newblock In \emph{Robotics and Automation, 1993. Proceedings., 1993 IEEE
  International Conference on}, pages 196--201. IEEE, 1993.

\bibitem[Bicchi(1994)]{BICCHI94}
Antonio Bicchi.
\newblock On the problem of decomposing grasp and manipulation forces in
  multiple whole-limb manipulation.
\newblock \emph{Robotics and Autonomous Systems}, 13\penalty0 (2):\penalty0
  127--147, 1994.

\bibitem[Bicchi(1995)]{BICCHI95}
Antonio Bicchi.
\newblock On the closure properties of robotic grasping.
\newblock \emph{The International Journal of Robotics Research}, 14\penalty0
  (4):\penalty0 319--334, 1995.

\bibitem[Cutkosky and Kao(1989)]{CUTKOSKY_COMPLIANCE}
Mark~R. Cutkosky and Imin Kao.
\newblock Computing and controlling the compliance of a robotic hand.
\newblock \emph{IEEE Transactions on Robotics and Automation}, 5\penalty0 (2),
  1989.

\bibitem[Eppstein(2008)]{EPPSTEIN08}
David Eppstein.
\newblock Recognizing partial cubes in quadratic time.
\newblock \emph{Journal of Graph Algorithms and Applications}, 15\penalty0 (2),
  2008.

\bibitem[Ferrari and Canny(1992)]{FERRARI92}
C.~Ferrari and J.~Canny.
\newblock Planning optimal grasps.
\newblock In \emph{IEEE International Conference on Robotics and Automation},
  pages 2290--2295, 1992.

\bibitem[Fukuda et~al.(1991)Fukuda, Saito, Tamura, and Tokuyama]{FUKUDA91}
Komei Fukuda, Shigemasa Saito, Akihisa Tamura, and Takeshi Tokuyama.
\newblock Bounding the number of k-faces in arrangements of hyperplanes.
\newblock \emph{Discrete Applied Mathematics}, 31, 1991.

\bibitem[Gleason et~al.(1980)Gleason, Kelly, and Greenwood]{GREATCIRCLES}
Andrew~M. Gleason, L.~M. Kelly, and R.~E. Greenwood.
\newblock Solutions: The twenty-fifth competition.
\newblock \emph{The William Lowell Putnam mathematical competition : problems
  and solutions :1938-1964}, 1980.

\bibitem[Liu(1999)]{YUNHUI99}
Yun-Hui Liu.
\newblock Qualitative test and force optimization of 3-d form closure grasps
  using linear programming.
\newblock \emph{IEEE Transactions on Robotics and Automation}, 15\penalty0 (1),
  1999.

\bibitem[Markenscoff et~al.(1990)Markenscoff, Ni, and
  Papadimitriou]{CHRISTOS90}
Xanthippi Markenscoff, Luqun Ni, and Christos~H. Papadimitriou.
\newblock The geometry of grasping.
\newblock \emph{The International Journal of Robotics Research}, 9\penalty0
  (1):\penalty0 61--74, 1990.

\bibitem[Mehlhorn and Michail(2005)]{MEHLHORN05}
Kurt Mehlhorn and Dimitrios Michail.
\newblock Implementing minimum cycle basis algorithms.
\newblock In Sotiris~E. Nikoletseas, editor, \emph{Experimental and Efficient
  Algorithms}, pages 32--43. Springer Berlin Heidelberg, 2005.

\bibitem[Pang and Trinkle(2000)]{ZAMM643}
J.-S. Pang and J.~Trinkle.
\newblock Stability characterizations of rigid body contact problems with
  coulomb friction.
\newblock \emph{ZAMM - Journal of Applied Mathematics and Mechanics /
  Zeitschrift f\"ur Angewandte Mathematik und Mechanik}, 80\penalty0 (10),
  2000.

\bibitem[Peshkin and Sanderson(1989)]{PESHKIN89}
Michael~A. Peshkin and Arthur~C. Sanderson.
\newblock Minimization of energy in quasi-static manipulation.
\newblock \emph{IEEE Transactions on Robotics and Automation}, 5\penalty0 (1),
  1989.

\bibitem[Prattichizzo et~al.(1997)Prattichizzo, Salisbury, and
  Bicchi]{PRATTICHIZZO97}
Domenico Prattichizzo, John~Kenneth Salisbury, and Antonio Bicchi.
\newblock Contact and grasp robustness measures: Analysis and experiments.
\newblock In \emph{Experimental Robotics IV}, pages 83--90. Springer Berlin
  Heidelberg, 1997.

\bibitem[Somov(1900)]{SOMOV1900}
P.~Somov.
\newblock \"uber gebiete von schraubengeschwindigkeiten eines starren k\"orpers
  bei verschiedener zahl von st\"utzfl\"achen.
\newblock \emph{Zeitschrift für Mathematik und Physik}, 45, 1900.

\end{thebibliography}

\end{document}